\documentclass[10pt,twocolumn,letterpaper]{article}

\usepackage{wacv}              
\usepackage[accsupp]{axessibility}  
\definecolor{wacvblue}{rgb}{0.21,0.49,0.74}
\usepackage[pagebackref,breaklinks,colorlinks,allcolors=wacvblue]{hyperref}
\usepackage[linesnumbered,ruled,vlined]{algorithm2e}

\def\wacvPaperID{2882} 
\def\confName{WACV}
\def\confYear{2026}

\title{InteracTalker: Prompt-Based Human-Object Interaction with Co-Speech Gesture Generation}

\author{Sreehari Rajan\textsuperscript{*} \quad Kunal Bhosikar\textsuperscript{*} \quad Charu Sharma\\
Machine Learning Lab,\\
IIIT Hyderabad, India\\
{\tt\small \{sreehari.rajan, kunal.bhosikar\}@research.iiit.ac.in \quad charu.sharma@iiit.ac.in}
}

\begin{document}

\twocolumn[{
\maketitle
\begin{center}
    \captionsetup{type=figure}
    \includegraphics[width=0.9\linewidth]{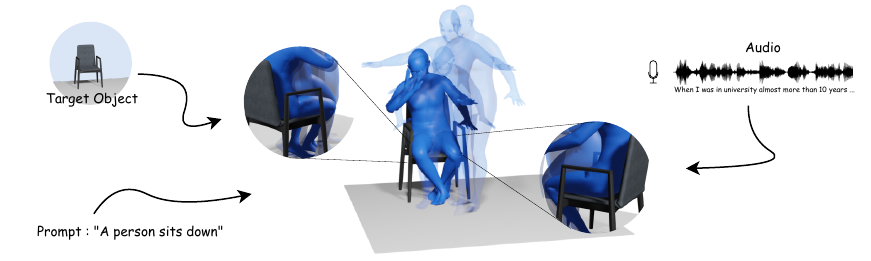}
    \captionof{figure}{InteracTalker generates realistic, full-body human motion, precisely integrating object interaction with expressive co-speech gestures, conditioned by speech audio, text prompts, and a target object.}
    \label{fig:teaser}
\end{center}
}]

\begin{abstract}
Generating realistic human motions that naturally respond to both spoken language and physical objects is crucial for interactive digital experiences. Current methods, however, address speech-driven gestures or object interactions independently, limiting real-world applicability due to a lack of integrated, comprehensive datasets. To overcome this, we introduce \textbf{InteracTalker}, a novel framework that seamlessly integrates prompt-based object-aware interactions with co-speech gesture generation. We achieve this by employing a multi-stage training process to learn a unified motion, speech, and prompt embedding space. To support this, we curate a rich human-object interaction dataset, formed by augmenting an existing text-to-motion dataset with detailed object interaction annotations. Our framework utilizes a Generalized Motion Adaptation Module that enables independent training, adapting to the corresponding motion condition, which is then dynamically combined during inference. To address the imbalance between heterogeneous conditioning signals, we propose an adaptive fusion strategy, which dynamically reweights the conditioning signals during diffusion sampling.
InteracTalker successfully unifies these previously separate tasks, outperforming prior methods in both co-speech gesture generation and object-interaction synthesis, outperforming gesture-focused diffusion methods, yielding highly realistic, object-aware full-body motions with enhanced realism, flexibility, and control.(\href{https://sreeharirajan.github.io/projects/InteracTalker/}{https://sreeharirajan.github.io/projects/InteracTalker/})
\end{abstract}
    
\section{Introduction}
\label{introduction}

Generating realistic, scene-aware human motions is critical for applications across computer graphics, robotics, and virtual environments, from gaming to embodied AI. While significant progress has been made in generating animations from text~\cite{tevet2023human, zhang2022motiondiffusetextdrivenhumanmotion} or speech~\cite{yi2023generatingholistic3dhuman, peng2024t3mtextguided3d}, these methods, including text-to-motion techniques~\cite{yi2024generating}, often focus on isolated aspects of human movement. However, these approaches typically lack the gestures that come with co-speech motion, resulting in motions that may not accurately reflect the realistic nature of human motion. Conversely, co-speech generation techniques~\cite{chhatre2024emotionalspeechdriven3dbody, Dan_ek_2023, ghorbani2022zeroeggszeroshotexamplebasedgesture, liu2022beat, 10.24963/ijcai.2023/650, zhu2023tamingdiffusionmodelsaudiodriven} can produce expressive gestures but frequently struggle to handle full-body object interactions. This fragmentation exists due to a lack of integrated datasets with multimodal annotations~\cite{guo2022generating, 9009460}.

\begin{figure}[t]
  \centering
  \includegraphics[width=0.9\linewidth]{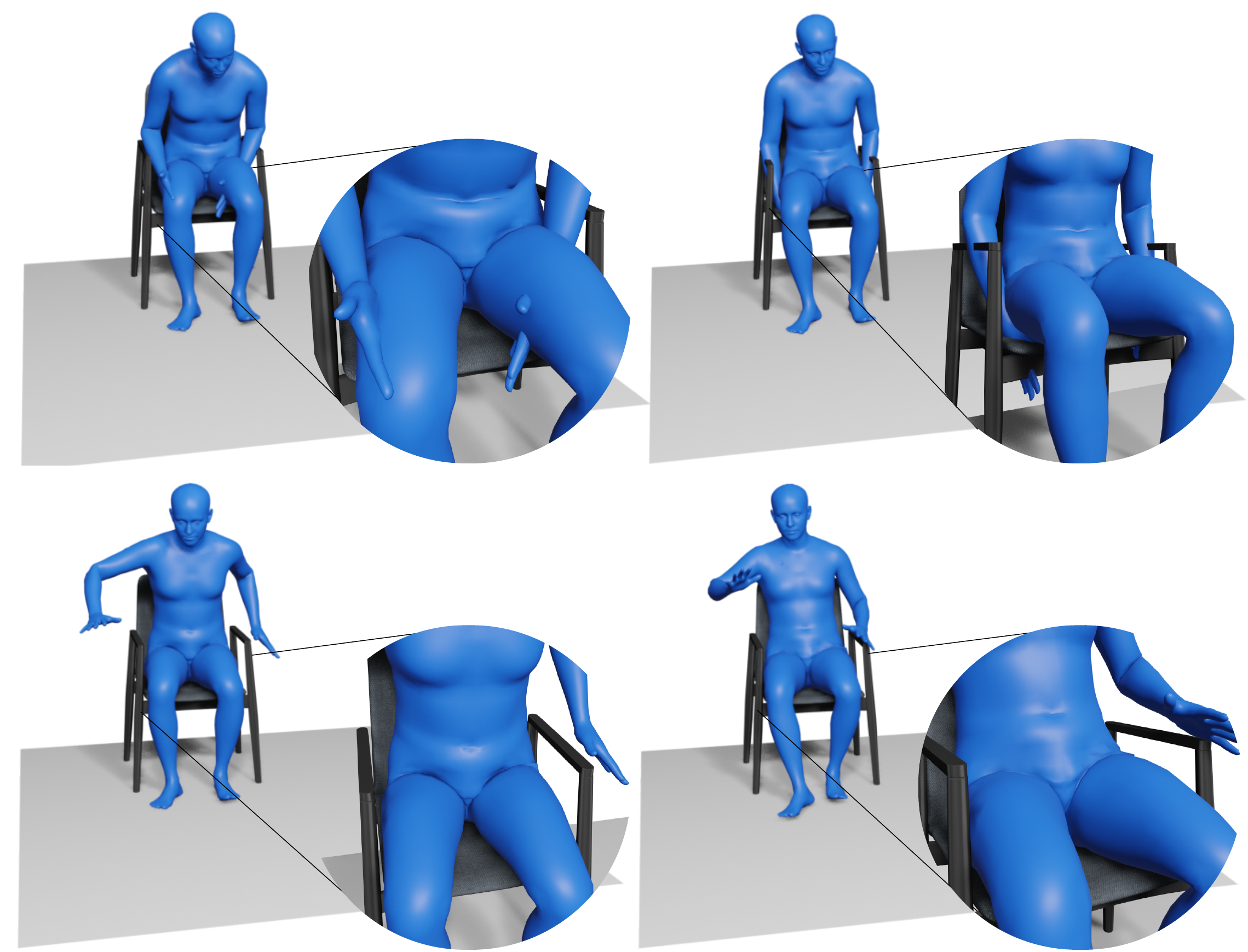}
  \caption{Comparison of Naive Motion Concatenation vs. InteracTalker. The top row shows a naive approach with severe artifacts like self- and object-penetration. The bottom row demonstrates our InteracTalker framework's coherent, artifact-free motion by jointly considering all signals.}
\label{fig: naive_concatenation_compare}
\end{figure}

A seemingly intuitive approach to combine these capabilities would be to naively concatenate independently generated co-speech (upper-body) and object-interaction (lower-body) motions. However, as illustrated in Figure~\ref{fig: naive_concatenation_compare} (top row), this fails. The motions lack mutual awareness, leading to undesirable self-penetrations and object penetrations. This highlights the critical need for a more integrated solution that truly understands the interplay between speech, body, and environment, Figure~\ref{fig: naive_concatenation_compare} (bottom row).

To overcome these limitations, we introduce \textbf{InteracTalker}, a novel framework that is the first one to seamlessly integrate text-conditioned object-aware motion generation with speech-driven co-speech gestures (Figure~\ref{fig:teaser}). This bridges a critical gap between isolated full-body motion synthesis and context-rich object interactions. 

Our key contributions are: 
(1) Unified Full-Body Motion Generation: InteracTalker unifies full-body co-speech gesture generation with prompt-based human-object interaction through a multi-stage training process to align motion, speech, and text prompts within a shared embedding space. To support this, we curated a human-object interaction dataset by enriching an existing text-to-motion dataset with detailed object interaction annotations. (2) Modular Adaptation Modules: We introduce modular adaptation modules specifically for object-driven interaction and co-speech-driven gesture. These are further extended into a Generalized Motion Adaptation Module, a flexible framework that allows for the integration of diverse conditioning signals without requiring computationally expensive retraining of the entire system. (3) Adaptive Fusion Strategy: A strategy to reweight the conditioning signals at each time step, ensuring a balanced fusion of heterogeneous conditioning signals. (4) Efficient and Robust Motion Generation: Our method eliminates the need for expensive test-time optimizations, producing significantly more realistic motions in considerably less time than existing methods. A key advantage of our Generalized Motion Adaptation Module is its ability to generate motion irrespective of the presence or absence of specific conditioning signals, all while maintaining a constant test-time. 

Extensive experiments demonstrate that InteracTalker not only effectively integrates these tasks but also outperforms state-of-the-art methods that tackle them individually, achieving greater realism, flexibility, and control in generating plausible, scene-aware full-body motions. In summary, InteracTalker represents a significant step towards unified and truly realistic human motion generation, paving the way for more immersive and lifelike digital characters.
\section{Related Work}
\label{related_work}

Human motion generation and human-object interaction have advanced significantly, focusing on realistic, context-aware motions for virtual humans~\cite{zhu2023humanmotiongenerationsurvey, 10.1145/566570.566607, taheri2023goalgenerating4dwholebody}. Early kinematic or physics-based models lacked context awareness for complex, real-world scenarios. Our work builds on these foundations by integrating multi-modal control.

\textbf{Text-Conditioned Human Motion Generation.} Text-based motion generation has progressed notably, enabling intuitive and versatile synthesis~\cite{zhang2022motiondiffusetextdrivenhumanmotion, tevet2023human, petrovich2024multitracktimelinecontroltextdriven}. TeSMo~\cite{yi2024generating} aligns motions with scene geometry from natural language. However, these methods often focus on abstract actions or predefined interactions, lacking detailed co-speech gestures or fine-grained object manipulations.

\textbf{Co-Speech Gesture Generation.} Another key area is expressive co-speech gesture generation synchronized with audio. Early work captured individual styles~\cite{ginosar2019learning} or leveraged trimodal inputs~\cite{yoon2020speech}. More recent advancements include enhancing coherence~\cite{liu2022learning}, diverse synthesis~\cite{liu2022disco}, and using diffusion models~\cite{yang2023diffusestylegesture}. BEAT~\cite{liu2022beat} introduced a large-scale dataset, while MambaTalk~\cite{xu2025mambatalkefficientholisticgesture} focuses on efficient holistic gesture synthesis. Freetalker~\cite{yang2024freetalkercontrollablespeechtextdriven} also uses diffusion models to generate gestures conditioned on both speech and text, demonstrating fine-grained control. While Freetalker~\cite{yang2024freetalkercontrollablespeechtextdriven} advances multi-conditional gesture synthesis, it focuses on upper-body gestures and overlooks the complex challenge of full-body interaction with objects, a limitation shared by most co-speech methods~\cite{yang2023diffusestylegesture}.

\textbf{Scene-Aware Human Motion and Broader Interactions.} Generating motions aware of and interacting with the environment or other agents is growing. TeSMo~\cite{yi2024generating} and Zhao et al.~\cite{zhao2023synthesizing} consider scene geometry but lack precise object interaction modeling. InterDance~\cite{li2024interdancereactive3ddancegeneration} explores reactive 3D dance duets, highlighting multi-agent synchronized motion complexity. While some frameworks integrate co-speech and full-body control~\cite{chen2024enabling}, they still often miss explicit human-object interaction, a critical limitation in cluttered environments. Other approaches align multimodal inputs for conversational gestures~\cite{yi2023generating}, learn 3D gestures from video~\cite{habibie2021learning}, or use expressive masked audio modeling like EMAGE~\cite{liu2024emage}.

\textbf{Unifying Co-Speech Gestures and Object Interactions.} Existing research typically addresses text-driven motion, co-speech gestures, or scene-aware human motion in isolation. Methods often excel in one area but rarely combine both simultaneously with fine-grained full-body control. For example, while diffusion models have been used for text and speech-driven gesture generation, they have not been successfully adapted to include object-aware interactions. This gap, integrating co-speech gestures with complex, dynamic human-object interactions within a unified framework, motivates InteracTalker, our novel framework designed for plausible, object-constrained, and co-speech full-body motions.

\begin{figure}[t]
    \centering
    \includegraphics[width=0.9\linewidth]{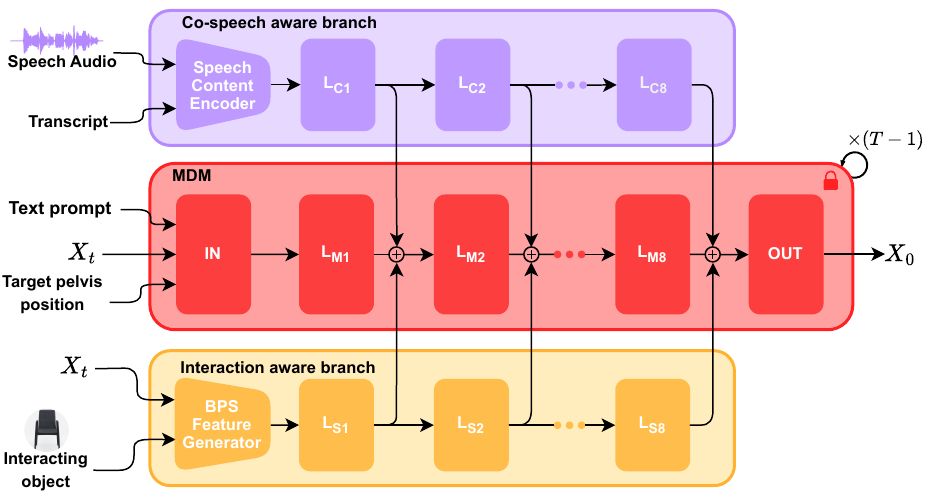}
    \caption{InteracTalker's architecture involves pre-training a Motion Diffusion Module (MDM), then fine-tuning Co-speech Gesture and Interaction-Aware branches; at inference, both branches condition the frozen MDM for holistic motion generation.}
    \label{fig:architecture}
\end{figure}
\section{Methodology}
\label{method}

\subsection{Overview}
Given speech audio, its corresponding transcript, a descriptive text prompt, and a specified target object, our objective is to generate realistic full-body human motions. These motions should simultaneously exhibit expressive co-speech gestures that capture the rhythm and semantics of the speech content, while also performing object-aware interactions guided by the text prompt and the target object's properties. The generated full-body motion will seamlessly integrate stylistic elements derived from both speech and text inputs.

Our InteracTalker architecture, illustrated in Figure~\ref{fig:architecture}, comprises three major conceptual components: (1) Motion Diffusion Module (MDM): This core module, based on the Motion Diffusion Model~\cite{tevet2023human}, is responsible for generating general, interaction-agnostic motion based solely on the text prompt. (2) Interaction-Aware Adaptation Branch: This component incorporates human-object interaction conditions, modifying the MDM's output to ensure realistic and physically plausible object interactions. (3) Co-Speech Gesture Adaptation Branch: This component similarly injects co-speech conditions into the MDM, enabling the generation of expressive gestures synchronized with speech.

By leveraging both adaptation branches, the generated motion comprehensively includes both realistic co-speech gestures and full-body object interactions. A multi-stage fine-tuning approach is employed to train these specialized branches.

\subsection{Human Motion Diffusion Model (MDM)}
Our foundational Motion Diffusion Model (MDM) generates motions by iteratively denoising a temporal sequence of \(N\) poses. During training, the model learns to reverse a forward diffusion process. Specifically, the denoising model predicts the denoised motion \(X_0\) from a noisy input motion \(X_t\). The training largely follows the standard procedure for denoising diffusion models~\cite{ho2020denoising}: a motion is sampled, noise is added to obtain \(X_t\), and the denoising model predicts the original motion, optimizing a reconstruction loss. During this denoising process, the model can be conditioned by various signals, such as a text prompt.

\begin{algorithm}[t]
    \SetAlgoLined
    \KwIn{Noisy input motion \(\mathbf{X}_t\), condition \(\mathbf{C_k}\)}
    \KwOut{Condition adapted noisy motion \(\mathbf{X}_{t-1}\)}
    \(\mathbf{c}_{k0} \leftarrow \mathbf{E}_{k}(\mathbf{C_k})\); \\
    \(\mathbf{H}_{M0} \leftarrow \mathbf{IN}(\mathbf{X}_t)\); \\
    \(\mathbf{H}_{M1} \leftarrow \mathrm{L}_{M1}(\mathbf{H}_{M0})\)\;
    \(\mathbf{c}_{k1} \leftarrow \mathrm{L}_{k1}(\mathbf{c}_{k0})\)\;
    \For{\(j \leftarrow 2\) \KwTo \(8\)}{
        \(\mathbf{H}_{Mj} \leftarrow \mathrm{L}_{Mj}(\mathbf{c}_{kj-1}+\mathbf{H}_{Mj-1})\)\;
        \(\mathbf{c}_{kj} \leftarrow \mathrm{L}_{kj}(\mathbf{c}_{kj-1})\)\;
    }
    \(\mathbf{X}_{t-1} \leftarrow \mathbf{OUT}(H_{M8} + \mathbf{c}_{k8})\); \\
    \textbf{Return:} \(\mathbf{X}_{t-1}\)
    \caption{Modular Motion Adaptation}
    \label{algo: conditioning_adaptation_module}
\end{algorithm}

\begin{figure*}[t]
    \centering
    \includegraphics[width=0.9\linewidth,trim=0pt 10pt 0pt 0pt, clip]{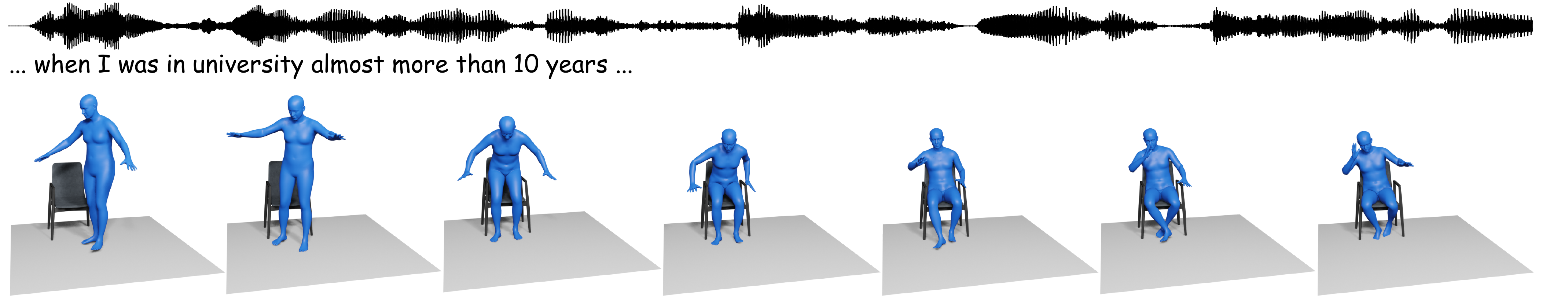}
    \includegraphics[width=0.9\linewidth,trim=0pt 10pt 0pt 80pt, clip]{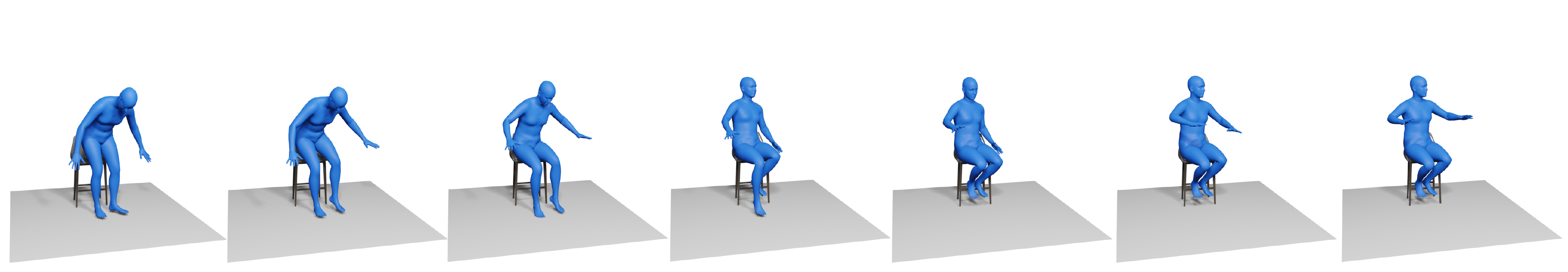}
    \includegraphics[width=0.9\linewidth,trim=0pt 20pt 0pt 60pt, clip]{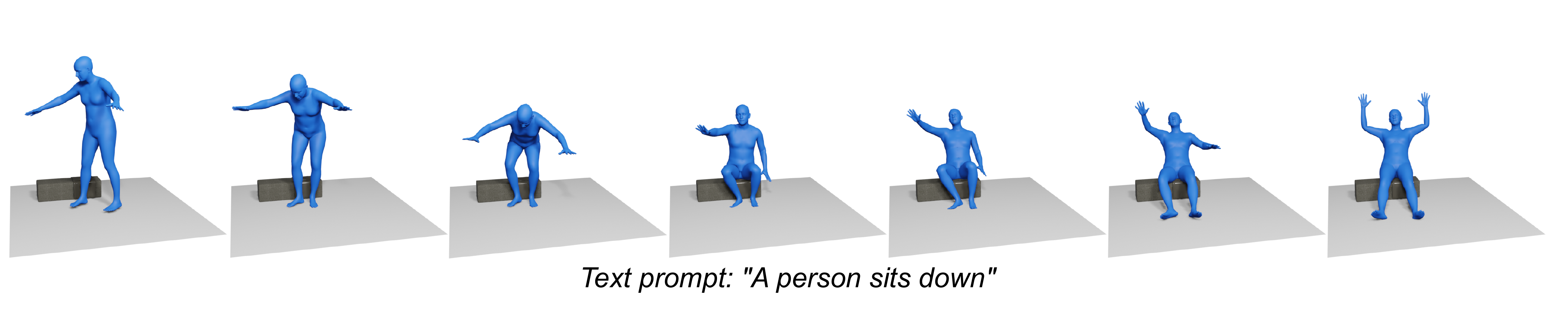}
    \caption{\textbf{InteracTalker:} Adaptive full-body motion with consistent co-speech gestures across various object heights (e.g., armchair, high stool, and Reebok step) for the same audio input.}
    \label{fig:qualitative_results}
\end{figure*}

Unlike some prior works that generate skeletons and then fit SMPL bodies to them~\cite{yi2024generating}, we directly utilize SMPL pose parameters. This approach enables the direct regression of SMPL parameters, effectively eliminating the need for computationally expensive post-processing optimization steps like those in~\cite{bogo2016keep}. Further details on our pose representation are provided in the Supplementary Materials.

The MDM architecture consists of linear input layers \((IN)\), eight transformer blocks \((L_{M1}, L_{M2}, ..., L_{M8})\) and output linear layers \((OUT)\). The diffusion model can be conditioned on a text prompt, which is encoded using CLIP~\cite{radford2021learning}, specifying a desired motion style and, optionally, a final goal pelvis position and orientation. We utilized a combination of the HumanML3D~\cite{guo2022generating} and SAMP~\cite{hassan2021stochastic} datasets to pre-train our base MDM.

\subsection{Generalized Motion Adaptation Modules}
The base MDM generates human motions without explicit environmental or expressive constraints. To significantly enhance the realism of these motions, particularly in complex scenarios, they must be aware of various contextual factors, such as their surroundings and co-speech cues. This awareness is achieved through our novel Generalized Motion Adaptation Modules. 

These modules share a common architectural design: an encoder \(E_k\) processes corresponding input constraints \(C_k\) (e.g., scene details, object geometry, co-speech audio). The encoded features are then passed through eight dedicated transformer blocks \((L_{k1}, L_{k2}, ..., L_{k8})\). The activations from each of these adaptation blocks serve as conditioning signals, which are then added to the corresponding transformer blocks within the MDM to generate adapted motions. This is shown in Algorithm~\ref{algo: conditioning_adaptation_module}. 

A key advantage of this modular design is the ability to train each adaptation module separately with the frozen base MDM. This multi-stage training process eliminates the need for a single, exhaustively annotated dataset encompassing all possible constraints, thereby improving data efficiency and scalability. 

\begin{figure}[t]
  \centering
  \includegraphics[width=0.9\linewidth]{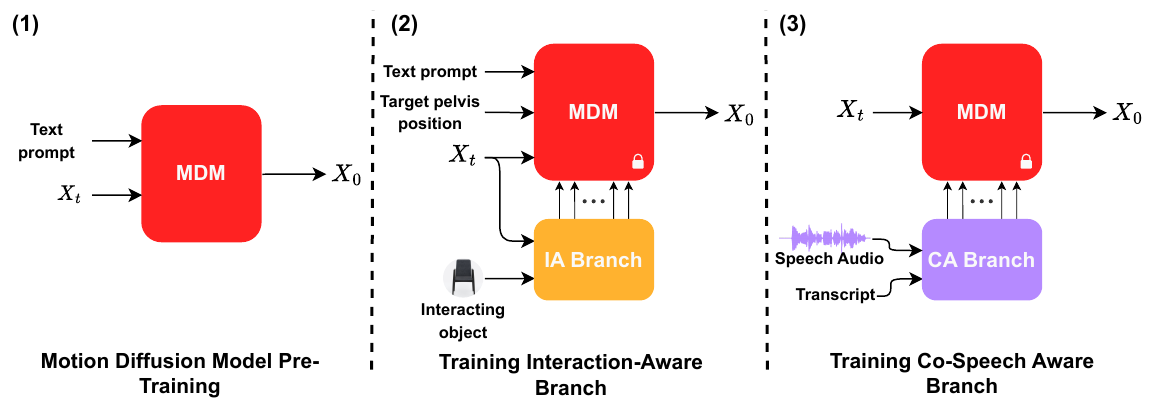}
  \caption{\textbf{InteracTalker's three-stage training pipeline:} (1) pre-training the base MDM, (2) fine-tuning the Interaction-Aware (IA) branch, and (3) fine-tuning the Co-speech Aware (CA) branch, all by injecting their signals into the frozen MDM.}
\label{fig: Training pipeline}
\end{figure}

\begin{figure*}[t]
  \centering
\includegraphics[width=0.9\linewidth,trim=0pt 0pt 0pt 0pt, clip]{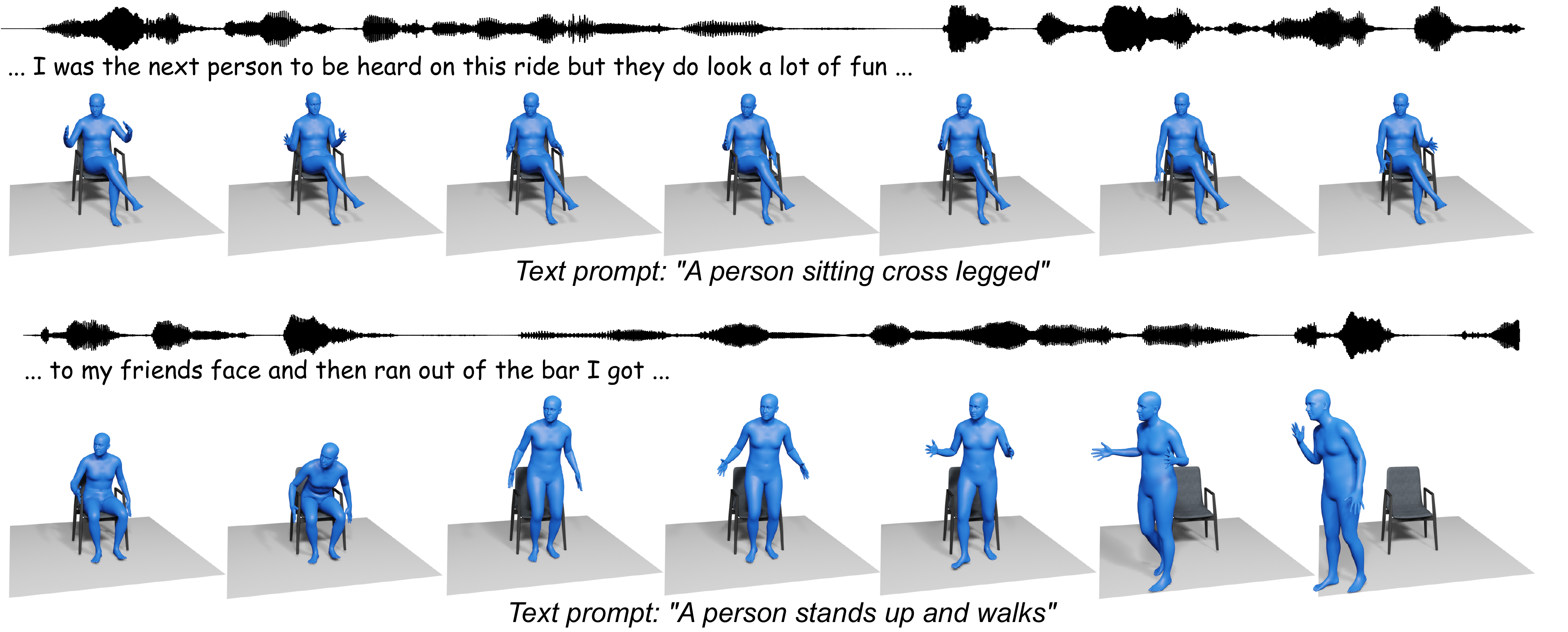}
 \caption{\textbf{InteracTalker:} Qualitative results showcasing diverse, object-aware motions with co-speech gestures, generated from varying text prompts for a single object.}
\label{fig:qualitative_diverse}
\end{figure*}

This modular adaptation framework is highly extensible and reusable across a variety of downstream tasks. It allows for the seamless integration of new conditioning modules without retraining of the entire system, effectively decoupling the main diffusion model from task-specific adaptations. Furthermore, multiple adaptation modules can be combined during inference to generate highly complex and multi-conditioned motions. We achieve this through an adaptive fusion strategy (refer to Section~\ref{subsubsec:afs}) that operates during the diffusion sampling process to balance these heterogeneous signals. Instead of using fixed weights, we dynamically reweight the contributions of each branch to ensure a coherent and plausible final motion. During inference, the guidance of each module is integrated as follows:
\begin{equation}
\label{guidance_eq}
    \mathcal{X}_t =  \mathcal{X}_{uncond} + \sum_{i=1}^n \lambda_{cond_i} \mathcal{X}_{cond_i} ,
\end{equation}
where \(\mathcal{X}_{uncond}\) is the unconditioned prediction (obtained by setting all conditioning signals to zero tensors), and \(\mathcal{X}_{cond_i}\) represents the guided prediction from individual conditioning \(cond_i\) (obtained by subtracting \(\mathcal{X}_{uncond}\) from the prediction when only \(cond_i\) is active, and all other conditionings are set to zero). 

\subsubsection{Object-Driven Interaction Motion Generation}
\label{obj_int_method}
The interaction-aware branch is specifically designed to enable realistic human-object interactions. It comprises a BPS (Basis Point Sets) feature generator (Figure 1b in Supplementary Materials) and eight transformer blocks. At each denoising step \(t\), the inputs to this branch are the current noisy body pose \(X_t\), and the target interaction object.

The BPS feature generator utilizes Basis Point Sets~\cite{prokudin2019efficient} to robustly represent object geometry and derive human-object interaction features~\cite{yi2024generating}. Object geometry features are computed as the minimum distance from each point in the BPS to the object surface. To generate human-object interaction features, we calculate the minimum distance from each BPS point to the 3D locations of the 22 SMPL joints from \(X_t\). These two sets of features are then merged using a Multi-Layer Perceptron (MLP) to obtain a learned combined feature vector, which serves as the input to the first transformer block \((L_{S1})\) of this branch.

\textbf{Interaction Motion Data:} To train this branch, we curated a dataset of interaction motions, specifically ``sitting from standing position", ``sitting", ``standing from sitting position", ``walk to sit", and ``from sitting position, stands and walks". These motions were extracted from the SAMP dataset~\cite{hassan2021stochastic}. To obtain corresponding motion-aligned 3D objects, we selected chairs of varying heights from the 3D-FRONT dataset~\cite{fu20213d} and manually aligned them to the extracted motions. We leveraged the textual descriptions of each extracted motion as annotated by~\cite{yi2024generating}.

\begin{table}[t]
    \scriptsize
    \centering
    \setlength{\tabcolsep}{4.3pt}
    \begin{tabular}{l|c|c|c|c|c}
        \toprule
        \textbf{Method} & \textbf{Motion} & \textbf{Natural} & \textbf{Object} & \textbf{Gesture} & \textbf{Fewer} \\
          & \textbf{Realism} & \textbf{Motion} & \textbf{Interaction} & \textbf{Quality} & \textbf{Penetrations} \\
         & \((1-5 \uparrow)\) & \((\% \uparrow)\) & \((\% \uparrow)\) & \((\% \uparrow)\) & \((\% \uparrow)\) \\
        \midrule
        Concat & 2.77 & 29.7\% & 20.3\% & 26.6\% & 20.3\% \\
        InteracTalker & \textbf{3.77} & \textbf{70.3\%} & \textbf{79.7\%} & \textbf{73.4\%} & \textbf{79.7\%} \\
        \bottomrule
    \end{tabular}
    \caption{User study comparing InteracTalker with a naive concatenation (Concat) method. Motion Realism is the average rating from 1 to 5. All other metrics show the percentage of participants who chose our method as superior in a two-alternative forced-choice question.}
    \label{tab:userstudy}
\end{table}

\textbf{Interaction Motion Training:} During training, the base MDM weights are frozen. The activations from each transformer block \((L_{S1}, L_{S2}, ..., L_{S8})\) of the interaction-aware branch are added to the corresponding activations of the MDM's transformer blocks \((L_{M1}, L_{M2}, ..., L_{M8})\), as outlined in Algorithm~\ref{algo: conditioning_adaptation_module}. This mechanism effectively injects the interaction-aware conditioning signals into the diffusion process. The training process is supervised by the following loss function:
\begin{equation}
\label{sw_loss}
    \mathcal{L}_{SW} =  \mathcal{L}_{rec} + \mathcal{L}_{pelvis} + \mathcal{L}_{contact} + \mathcal{L}_{collision} ,
\end{equation}
where,
\(\mathcal{L}_{rec} =  \left\| X_0 - x_0\right\|^{2}\) is the L2 reconstruction loss, \(\mathcal{L}_{pelvis}\) is L2-loss between the final pelvis pose of the ground truth and the denoised output, \(\mathcal{L}_{contact}\) and \(\mathcal{L}_{collision}\) are contact and collision losses, respectively, adapted from~\cite{yi2022human}.

Crucially, since only a subset of frames typically involves direct human-object interaction, we apply \(\mathcal{L}_{contact}\) and \(\mathcal{L}_{collision}\) selectively to a dynamically chosen subset of frames. For motions ending in a ``sitting position", the last few frames are selected for supervision. Similarly, for motions beginning from a ``sitting position", the initial few frames are chosen. This targeted supervision, unlike previous works such as~\cite{yi2024generating}, enables effective collision and penetration awareness during training, thereby eliminating the need for expensive test-time optimizations. 

\subsubsection{Co-Speech Driven Gestures Motion Generation}
The co-speech gesture branch is designed to generate expressive body gestures synchronized with spoken language. This branch consists of a speech content encoder (Figure 1a in Supplementary Materials) and eight transformer blocks. The speech content encoder takes audio and its corresponding transcript as input. These are processed separately to obtain distinct audio and text features. These features are then concatenated and passed through a linear layer to derive a joint audio-text feature vector, which subsequently feeds into the stack of transformer blocks \((L_{C1}, L_{C2}, ..., L_{C8})\).

\begin{table}[t]
    \scriptsize
  \centering
  \begin{tabular}{l|c|c|c}
    \toprule
    \textbf{Method}     & \textbf{FGD}\(\downarrow\)     & \textbf{BC}\(\uparrow\) & \textbf{Diversity}\(\uparrow\) \\
    \midrule
    S2G~\cite{ginosar2019learning} & 25.129 & 6.902 & 7.783     \\
    Trimodal~\cite{yoon2020speech} & 19.759 & 6.442 & 8.894 \\
    HA2G~\cite{liu2022learning} & 19.364 & 6.601 & 9.671 \\
    DisCo~\cite{liu2022disco} & 21.170 & 6.571 & 10.378 \\
    CaMN~\cite{liu2022beat} & 8.752 & 6.731 & 9.279 \\
    DiffStyleGesture~\cite{yang2023diffusestylegesture} & 10.137 & 6.891 & 11.075 \\
    Habibie \textit{et al}.~\cite{habibie2021learning} & 14.581 & 6.779 & 8.874 \\
    TalkShow~\cite{yi2023generating} & 7.313 & 6.783 & \underline{12.859} \\
    EMAGE~\cite{liu2024emage} & \underline{5.423} & 6.794 & \textbf{13.057} \\
    SynTalker~\cite{chen2024enabling} & 6.413 & \textbf{7.971} & 12.721 \\
    InteracTalker (Ours) & \textbf{1.017} & \underline{7.074} & 8.136\\
    \bottomrule
  \end{tabular}
  \caption{\textbf{Quantitative comparison of co-speech gesture generation on BEATX:} Performance metrics (FGD \(\times 10^{-1}\), BC \(\times 10^{-1}\), Diversity) are reported, with best in bold and second-best underlined.}
  \label{tab:quantitative_beatx}
\end{table}

\textbf{Co-Speech Gesture Data:} For training our Co-Speech Gesture Adaptation Branch, we utilized the comprehensive BEATX-Standard dataset~\cite{liu2024emage}. This large-scale dataset is particularly well-suited for our task, as it comprises over 30 hours of high-quality co-speech motion data, meticulously synchronized with corresponding audio recordings and their accurate transcripts. Its rich collection of natural conversational gestures provides ample data for learning the intricate relationships between speech and body movements.

\textbf{Co-Speech Gesture Training:} Similar to the interaction-aware branch, the base MDM weights are frozen during the training of this branch. The activations from each transformer block \((L_{C1}, L_{C2}, ..., L_{C8})\) of the co-speech branch are added to the corresponding activations of the MDM's transformer blocks \((L_{M1}, L_{M2}, ..., L_{M8})\). This mechanism effectively injects the co-speech-aware conditioning signals into the diffusion model. The training of this branch is supervised solely by the reconstruction loss, \(\mathcal{L}_{rec}\).

\subsubsection{Adaptive Fusion Strategy}
\label{subsubsec:afs}
Direct fusion of these heterogeneous conditioning signals results in imbalance. During diffusion sampling, the relative relevance of the signals might change dynamically; therefore, fixed or manually adjusted guidance weights are unable to address this.

\begin{figure}[t]
    \centering
    \includegraphics[width=0.9\linewidth,trim=0pt 0pt 0pt 0pt, clip]{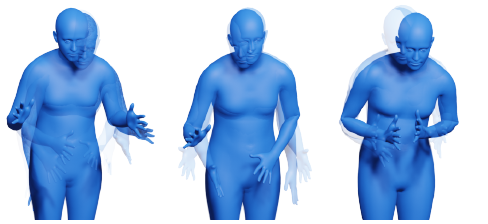}
    \caption{\textbf{Qualitative results of InteracTalker's Co-Speech Gesture Adaptation Branch:} expressive, diverse gestures precisely aligned with speech audio on the BEATX dataset.}
    \label{fig:qualitative_beatx}
\end{figure}

\begin{figure*}[t]
  \centering
  \includegraphics[width=0.9\linewidth,trim=0pt 0pt 0pt 20pt, clip]{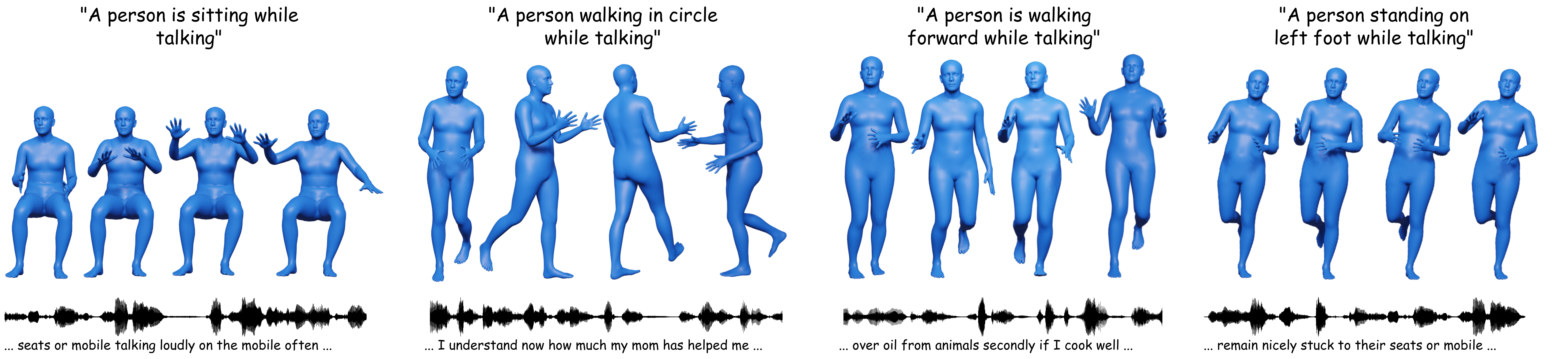}
 \caption{\textbf{Qualitative results of InteracTalker's Co-Speech Gestures Adaptation Branch:} diverse, natural, and expressive full-body gestures synchronized with various speech inputs in general settings.}
\label{fig:general_cospeech}
\end{figure*}

To overcome this, we introduced an adaptive fusion strategy during diffusion sampling, which involves updating these guidance weights, \(\lambda_{cond_i}\) in each timestep. First, to avoid scale imbalance, we normalized the co-speech residual, \(\mathcal{X}_{cospeech}\), relative to the interaction-aware residual, \(\mathcal{X}_{int}\). Next, to adaptively learn the guidance weights, \(\lambda_{cospeech}\) and \(\lambda_{int}\), we employ pseudo-supervision using single-conditioned predictions. Specifically, we compute motion predicted when only the interaction-aware branch is active (\(X_0^{int}\)) and when only the cospeech-aware branch is active (\(X_0^{cospeech}\)). These serve as anchors that represent how motion would evolve if each conditioning signal were considered in isolation. During diffusion sampling, we encourage the fused prediction \(\mathcal{X}_t\) to stay consistent with both anchors by updating the guidance weights \(\lambda_{cond_i}\) via gradient-based self-supervision, without requiring ground-truth. 
At each timestep, \(t\), we then measure how far the fused prediction \(\mathcal{X}_t\) deviates from each anchor by computing the two losses given: 
\begin{equation}
\label{af_int}
    \mathcal{L}_{AF_{int}} = \left\|  \mathcal{X}_t -  \mathcal{X}_0^{int}\right\|^{2},
\end{equation}
\begin{equation}
\label{af_cospeech}
    \mathcal{L}_{AF_{cospeech}} = \left\|  \mathcal{X}_t -  \mathcal{X}_0^{cospeech}\right\|^{2},
\end{equation}

These losses serve as soft alignment signals: if the fused motion deviates too much from the anchor of one condition, the corresponding guidance weight is raised to amplify the effect of that branch. Conversely, if the alignment is already good, its guidance weight naturally decreases. We update \(\lambda_{cospeech}\) and \(\lambda_{int}\) using gradient-based online updates at every denoising step:
\begin{equation}
\label{af_cospeech_updaterule}
    \lambda_{cospeech_{t-1}} = \lambda_{cospeech_{t}} -  \frac{\partial \mathcal{L}_{AF_{cospeech}}}{\partial \lambda_{cospeech_{t}}}, 
\end{equation}
\begin{equation}
\label{af_int_updaterule}
    \lambda_{int_{t-1}} = \lambda_{int_{t}} -  \frac{\partial \mathcal{L}_{AF_{int}}}{\partial \lambda_{int_{t}}}, 
\end{equation}
where, \(\lambda_{cospeech_{t}}\) and \(\lambda_{int_{t}}\) are the guidance weights at timestep \(t\) and \(\lambda_{cospeech_{t-1}}\) and \(\lambda_{int_{t-1}}\) are the guidance weights at timestep \(t-1\).

This approach ensures that the final motion is not only physically plausible but also stylistically consistent, as the model dynamically prioritizes the most relevant conditioning signal at each step of the denoising process.
\section{Experiments}
\label{experiments}

Our training pipeline, illustrated in Figure~\ref{fig: Training pipeline}, follows a three-stage, modular approach to overcome the challenge of limited fully annotated data. We first pre-train the base Motion Diffusion Model (MDM), using the HumanML3D~\cite{guo2022generating} and SAMP~\cite{hassan2021stochastic} datasets. Subsequently, we train the Interaction-Aware (IA) Branch using a custom dataset derived from SAMP~\cite{hassan2021stochastic} and motion-aligned objects from the 3D-FRONT~\cite{fu20213d} dataset. Finally, the Co-speech Gesture Adaptation Branch is trained using the BEATX~\cite{liu2024emage} dataset. During inference, signals from both branches are synergistically injected into the frozen MDM to generate comprehensive, multi-conditioned motions. A detailed description of our experimental setup is provided in the Supplementary Materials.

\subsection{Holistic Evaluation: Interaction-Aware Co-Speech Gesture Generation}
As the first work to address the specific, complex task of generating co-speech gestures that simultaneously account for detailed object interactions, we showcase the qualitative capabilities in Figure~\ref{fig:qualitative_results} and Figure~\ref{fig:qualitative_diverse}, demonstrating the effectiveness of our approach in combining the conditioning effects of both co-speech dynamics and object interactions. Table~\ref{tab:userstudy} is a user study that compares our full InteracTalker framework with a naive concatenation (Concat) approach, which combines independently generated upper-body co-speech gestures and lower-body object-interaction motions.

\begin{table}[t]
    \scriptsize
  \centering
  \begin{tabular}{l|c|c|c|c|c}
    \toprule
    & \multicolumn{3}{c|}{\textbf{Goal-reaching error}\((\downarrow)\)} &  \multicolumn{2}{c}{\textbf{Obj. penetration}\((\downarrow)\)} \\
    \cmidrule(){2-4}\cmidrule(){5-6}
    \textbf{Method}     & \textbf{Pos.} & \textbf{Height} & \textbf{Orient.} & \textbf{Value} & \textbf{Ratio}\\
    \midrule
    DIMOS~\cite{zhao2023synthesizing} & 0.2020 & 0.1283 & 0.4731 & 0.0193 & 0.1076\\
    TeSMo~\cite{yi2024generating} & \underline{0.1445} & \textbf{0.0120} & \underline{0.2410} &  \underline{0.0043} & \textbf{0.0611}\\
    InteracTalker & \textbf{0.1247} & \underline{0.0388} & \textbf{0.0965} & \textbf{0.0010} & \underline{0.0619}\\
    \bottomrule
  \end{tabular}
  \caption{\textbf{Quantitative performance on human-object interaction:} InteracTalker (Interaction-Aware branch) compared to SOTA on the SAMP sitting dataset. Metrics include Goal-reaching Error (Pos, Height, Orient) and Object Penetration (Value, Ratio). The best values are in \textbf{bold} and the second-best are \underline{underlined}.}
  \label{tab:quantitative_samp}
\end{table}

In Figure~\ref{fig:qualitative_results}, we present results across three chairs of varying heights. These examples highlight InteracTalker's ability to generate co-speech gestures that not only align closely with the input speech audio's rhythm and semantics but also appropriately adjust full-body movements and poses based on the affordances and constraints of the target object's geometry. Furthermore, Figure~\ref{fig:qualitative_diverse} illustrates the adaptability and versatility of our method by showcasing various human-object interactions with the same object, all while maintaining natural co-speech gestures. This demonstrates the rich diversity of motions our integrated framework can achieve.

User study was conducted with 64 non-expert participants, who compared paired videos with speech audio generated by our method and the naive concatenation approach. Participants rated motion quality across several aspects, and results show a clear preference for InteracTalker, with significantly higher realism scores and fewer artifacts such as self- and object-penetrations (see Table~\ref{tab:userstudy} and Supplementary Materials for details).

\subsection{Co-Speech Gesture Generation}
To evaluate the performance of our Co-speech Gesture Adaptation Branch independently, we compare our method against state-of-the-art speech-to-motion generation techniques. This isolated evaluation clearly assesses the effectiveness of our co-speech-aware component.

\begin{figure}[t]
    \centering
    \includegraphics[width=0.9\linewidth,trim=0pt 0pt 0pt 0pt, clip]{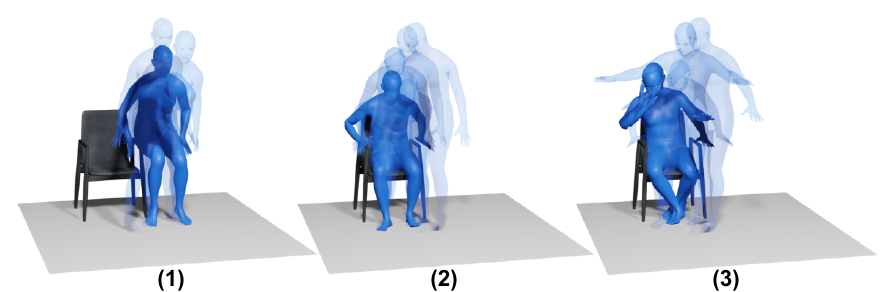}
    \caption{\textbf{Qualitative Ablation:} Incremental realism shown using a consistent prompt and object. Motion progresses from (1) interaction-agnostic baseline, (2) object-aware motion (IA Branch), to (3) fully integrated motion with co-speech gestures (IA + CA Branch).}
    \label{fig:qualitative_ablation}
\end{figure}

Quantitative comparisons in Table~\ref{tab:quantitative_beatx}, which utilize established metrics like Frechet Gesture Distance (FGD)~\cite{yoon2020speech}, Diversity~\cite{li2021audio2gestures}, and Beat Consistency (BC)~\cite{li2021ai}, indicate that our approach achieves comparable, and in several cases superior, performance relative to current state-of-the-art methods. Qualitative results, further illustrating the diverse and natural co-speech gestures generated by our method, are presented in Figure~\ref{fig:qualitative_beatx}.

Additionally, to assess the generation of full-body motions conditioned on both speech and text prompts, we conducted a qualitative experiment with diverse text prompts and speeches, presented in Figure~\ref{fig:general_cospeech}. This shows that our method can also generate co-speech gestures that closely align with the input speech, along with general motions given by the text prompts.

\subsection{Human-Object Interaction Generation}
We evaluate the Interaction-Aware Adaptation Branch independently to directly compare its performance against state-of-the-art human-object interaction methods. Quantitative and qualitative results are presented in Table~\ref{tab:quantitative_samp} and the Supplementary Materials, respectively.

Table~\ref{tab:quantitative_samp} critically highlights our method's superior performance in goal-reaching tasks without requiring any test-time guidance or optimization, a significant advantage over previous state-of-the-art approaches. Specifically, our method consistently achieves lower object penetration values and ratios, indicating substantially more physically plausible and collision-free interactions. We define the object penetration value as the average Signed Distance Function (SDF) value across all penetrated body vertices in the generated motions, while the object penetration ratio represents the fraction of generated poses exhibiting penetration, calculated over all frames in the motion sequence.

\subsection{Ablation Study}
\label{subsec:ablation}

To understand the individual contributions of each adaptation branch, we conducted a qualitative ablation study using a consistent text prompt (``A person sits down'') and target object, as shown in Figure~\ref{fig:qualitative_ablation}. Without any adaptation branches, the motion is interaction-agnostic (Figure~\ref{fig:qualitative_ablation}, left), leading the human body to simply ``sit'', disregarding the chair.

\begin{figure}[t]
    \centering
    \includegraphics[width=0.9\linewidth]{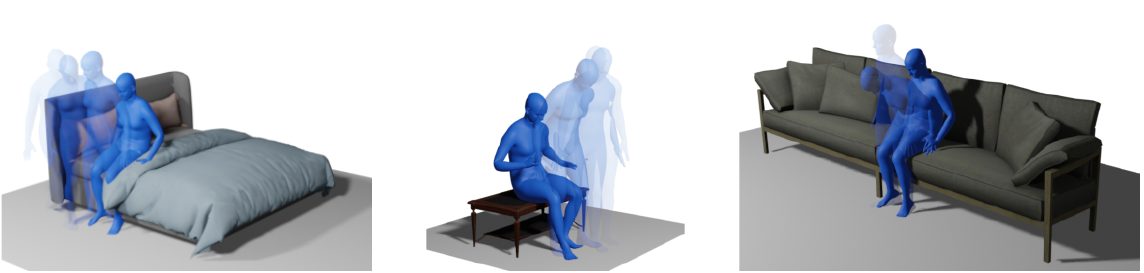}
    \caption{\textbf{Qualitative results:} InteracTalker demonstrates robust generalization to novel, unseen object geometries, showing successful navigation and appropriate physical engagement.}
    \label{fig:qualitative_generalized}
\end{figure}

Interaction-Aware Motion: By activating only the Interaction-Aware branch (Figure~\ref{fig:qualitative_ablation}, middle), the generated motion becomes object-aware. Instead of sitting directly, the human character appropriately takes steps towards the chair and then performs a plausible sitting motion on it.

Interaction-Aware + Co-Speech Motion: Finally, by incorporating both the Interaction-Aware and Co-speech Gesture branches (Figure~\ref{fig:qualitative_ablation}, right), the character not only performs the object-aware sitting action but also exhibits natural, synchronized co-speech gestures, demonstrating the full integration capability of InteracTalker.

\subsection{Generalization to Unseen Objects}
To assess the generalizability of our method beyond its training data, we performed human-object interaction experiments with unseen and complex object types, presented in Figure~\ref{fig:qualitative_generalized}. Although the Interaction-aware branch was primarily trained on chairs of varying heights, this experiment aims to confirm its robustness and adaptability to novel object geometries. As illustrated in Figure~\ref{fig:qualitative_generalized}, our method successfully understands the complex nature of these unseen objects and is able to navigate the character to the most suitable area for interaction. This qualitatively proves the strong adaptability of InteracTalker with complex, unseen object types for interaction scenarios.

\section{Conclusion}
\label{conclusion}
We introduced InteracTalker, a novel framework unifying co-speech gestures and prompt-based human-object interactions, which addresses isolated approaches in motion generation. Our framework, which leverages a multi-stage training process, modular adaptation modules, and an adaptive fusion strategy, successfully unifies and outperforms prior methods in each subtask, yielding realistic, object-aware full-body motions with gestures, offering superior realism, flexibility, and control.

\textbf{Limitations and Future Work:} While InteracTalker marks a significant advance, it has some limitations. The current framework focuses on a single character, whereas real conversations often involve multiple characters. Extending to multi-character is an exciting future direction. Our approach also does not explicitly account for stylistic variations like speaker identity or emotion, which would result in more customized motion. Additionally, addressing the scarcity of diverse human-object interaction datasets, extending to complex scene-aware motions, and automating target pelvis position inference from text prompts remains crucial for broader applicability.
\section*{Acknowledgments}
\label{acknowledgement}

The authors gratefully acknowledge the support of the Science and Engineering Research Board (SERB), Government of India, for providing GPU computing resources that enabled this research.

\appendix

\section*{Appendix}
In the Appendix, we provide the following:
\begin{itemize}
    \item 3d human motion representation details in Appendix~\ref{pose_rep_appendix}.
    \item detailed discussion on adaptation module encoders in Appendix~\ref{adaptation_module_encoder}.
    \item comprehensive implementation details and experimental setup in Appendix~\ref{exp_setup}.
    \item qualitative examples of physically plausible human-object interaction by InteracTalker in Appendix~\ref{qualitative_appendix}.
    \item an expanded review of the conducted user study in Appendix~\ref{user_study}.
\end{itemize}

\section{3D Human Motion Representation}
\label{pose_rep_appendix}
Our 3D human motion is represented using SMPL pose parameters, a standard practice in recent work. The orientation of each of the 22 joints (including the root joint) is encoded using 6D rotations~\cite{zhou2019continuity} to avoid gimbal lock and ensure continuity. Instead of using absolute global translations, we represent the global body movement as the difference between consecutive frames~\cite{athanasiou2024motionfix,holden2016deep}. This approach helps to generate smoother motions by focusing on local displacement rather than absolute position, which can suffer from drift. Consequently, each motion frame is represented as a 135-dimensional vector, comprising the 6D rotations for all 22 joints and a 3D vector for the consecutive global translation. A full motion is a sequence of these 135-dimensional vectors. Before training, these features are normalized using the mean and variance of the training set, a standard procedure to stabilize the learning process.

\section{Adaptation Module Encoders}
\label{adaptation_module_encoder}

\begin{figure}[h!]
    \centering
    \includegraphics[width=\linewidth]{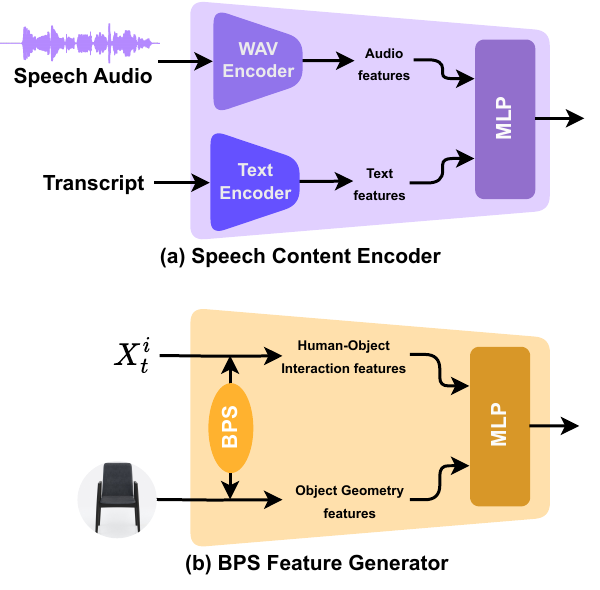}
    \caption{Architecture of InteracTalker's Adaptation Encoders. (a) The Speech Content Encoder extracts joint audio-text features from speech audio and a transcript for co-speech gesture generation. (b) The BPS Feature Generator derives object geometry and human-object interaction features from the body pose and a target object to guide object-aware motion.}
    \label{fig:module_encoders}
\end{figure}

This section provides a detailed overview of the two primary encoders used to generate conditioning signals for our adaptation modules. The architecture of these encoders is illustrated in Figure~\ref{fig:module_encoders}.

As shown in Figure~\ref{fig:module_encoders}(a), the Speech Content Encoder is designed to process the multimodal inputs required for co-speech gesture generation. It takes both speech audio and its corresponding transcript, processes them to extract distinct features, and then concatenates these to produce a joint audio-text feature vector. This feature vector subsequently serves as the conditioning signal for the Co-speech Gesture Adaptation Branch.

Figure~\ref{fig:module_encoders}(b) details the BPS Feature Generator, which is responsible for deriving object-specific conditioning signals. This encoder takes as input the current noisy body pose and the geometry of the target object. It extracts two types of features: object geometry features from Basis Point Sets (BPS) and human-object interaction features. These are then merged to create a comprehensive feature vector that guides the Interaction-Aware Adaptation Branch, enabling our model to generate motions that are highly responsive to a target object's properties.

\section{Experimental Setup}
\label{exp_setup}

Our experimental setup follows a three-stage training process, leveraging a modular approach to build the InteracTalker framework.

First, the base Motion Diffusion Model (MDM) was pre-trained using the HumanML3D~\cite{guo2022generating} and SAMP~\cite{hassan2021stochastic} datasets, which do not contain object interactions. This training was performed for 500k steps with a batch size of 200. Following this, the MDM was frozen to serve as a stable motion prior for the subsequent training of our adaptation branches.

Next, the interaction-aware branch was trained for 110k steps with a batch size of 64. This stage utilized the SAMP~\cite{hassan2021stochastic} dataset, augmented with motion-aligned objects from the 3D-FRONT~\cite{fu20213d} dataset, along with their corresponding text prompts.

Finally, the co-speech aware branch was trained separately for 600k steps with a batch size of 64 using the BEATX~\cite{liu2024emage} dataset. This allowed it to specialize in the fine-grained dynamics of co-speech gestures. 

For all training stages, we used the AdamW optimizer with a learning rate of \(10^{-4}\). All experiments were conducted on a single NVIDIA GeForce RTX 2080 Ti GPU, with the total training time spanning approximately 5 days. The number of diffusion steps was set to 1000 for both training and inference.

During inference, the conditioning signals from both adaptation branches are synergistically injected into the pre-trained and frozen MDM, enabling the generation of comprehensive motions that account for both co-speech dynamics and object interactions simultaneously.

\section{Qualitative Results of InteracTalker}
\label{qualitative_appendix}

\begin{figure}[h!]
    \centering
    \includegraphics[width=\linewidth]{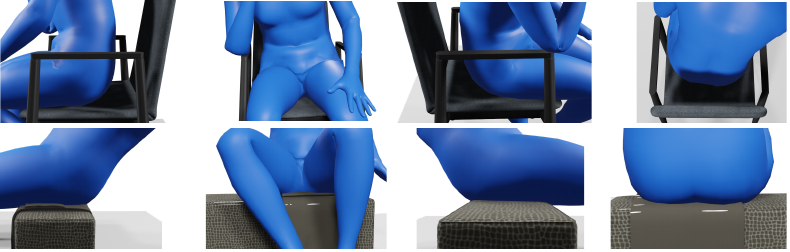}
    \caption{Qualitative Results: Physically Plausible Human-Object Interaction by InteracTalker. This figure demonstrates our method's superior ability to generate realistic human-object interactions. Observe how our approach consistently maintains necessary physical contact with the target object (e.g., foot on ground, hand on chair) while achieving minimal penetration, a key challenge in interaction synthesis. These results visually corroborate our low quantitative penetration metrics.}
    \label{fig:qualitative_pene_contact}
\end{figure}

This section provides a visual demonstration of our method's performance in generating realistic and physically plausible human-object interactions. Close-up views of the interaction regions in the generated motions are presented in Figure~\ref{fig:qualitative_pene_contact}, visually demonstrating that our approach effectively maintains the necessary physical contact for realistic interaction while rigorously minimizing penetration with the target object.

In addition to Figure~\ref{fig:qualitative_pene_contact} presented below, a supplementary video is provided to offer a dynamic view of our qualitative results. A `README' file is also included with the video, which details the specific content and examples shown.

\section{User Study}
\label{user_study}
\begin{figure*}[h!]
    \centering
    \includegraphics[width=\linewidth]{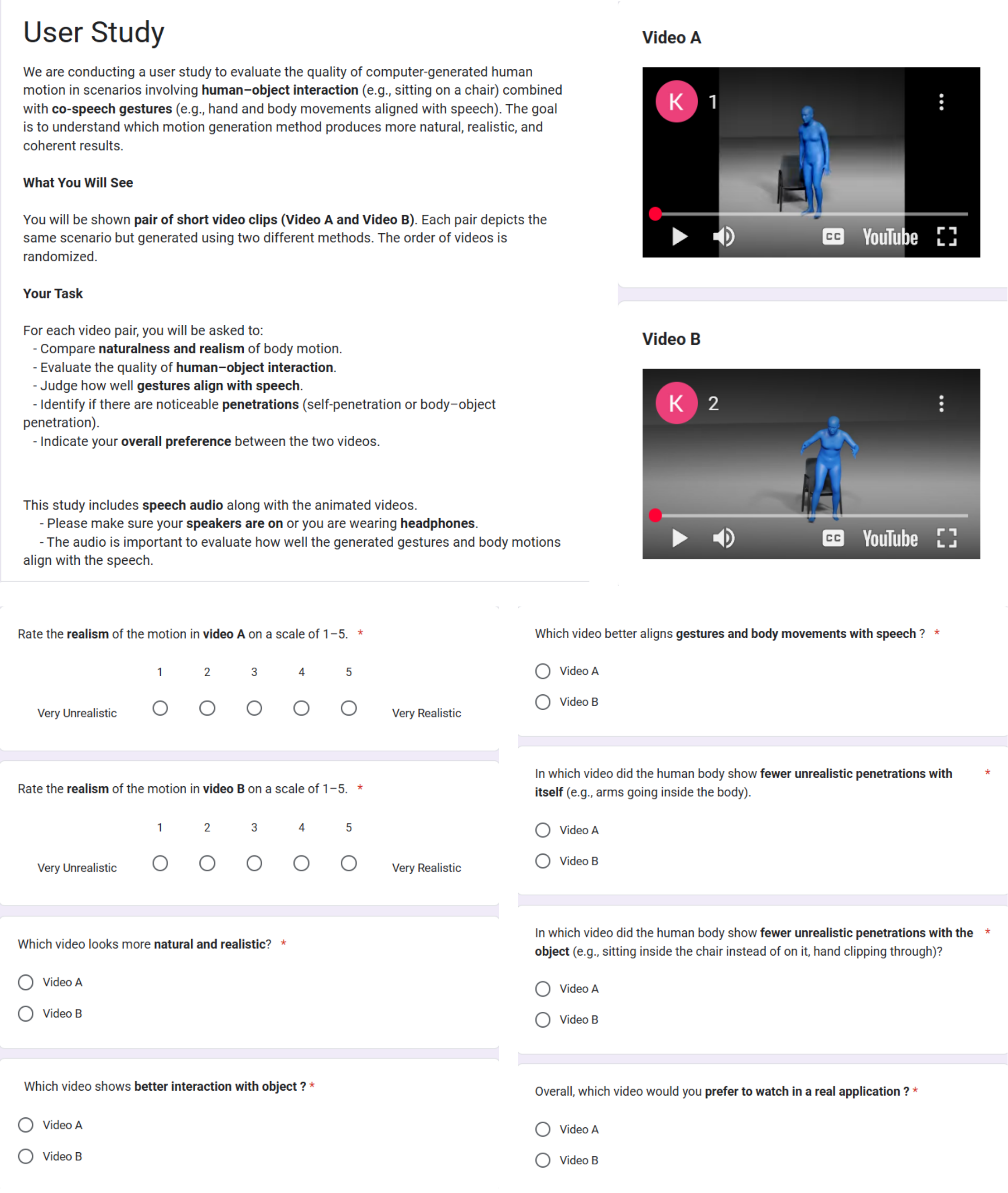}
    \caption{Interface used in conducting user study.}
    \label{fig:user_study}
\end{figure*}

As the first work to address the complex task of generating co-speech gestures that simultaneously account for detailed object interactions, we conducted a user study to assess the perceived plausibility and realism of our method. We compared our full InteracTalker framework with a naive concatenation (Concat) approach, which combines independently generated upper-body co-speech gestures and lower-body object-interaction motions. We recruited around 64 participants with minimal or no prior expertise in the domain to ensure an unbiased evaluation based on visual clarity and naturalness. Participants were presented with pairs of videos from both our method and the naive approach. For each pair, they answered questions about different aspects of motion quality. The full evaluation interface is provided in Figure~\ref {fig:user_study}. The user study results, demonstrate a strong preference for our InteracTalker framework (preferred 75\%). Our method achieved a significantly higher average rating for motion realism and was preferred by a large majority of participants across all other metrics. This confirms that a unified, integrated solution is critical for generating realistic motions, as the naive approach consistently produced motions with undesirable artifacts such as self- and object-penetrations.

{
    \small
    \bibliographystyle{ieeenat_fullname}
    \bibliography{main}
}

\end{document}